\documentclass[conference]{IEEEtran}
\IEEEoverridecommandlockouts

\usepackage{cite}
\usepackage{amsmath,amssymb,amsfonts}
\usepackage{algorithmic}
\usepackage{graphicx}
\usepackage{textcomp}
\usepackage{xcolor}

\usepackage{booktabs}
\usepackage{multirow}

\usepackage[percent]{overpic}

\def\BibTeX{{\rm B\kern-.05em{\sc i\kern-.025em b}\kern-.08em
    T\kern-.1667em\lower.7ex\hbox{E}\kern-.125emX}}
\begin{document}

\title{Lesion-Aware Attention for Mammography Segmentation with Heterogeneous Supervision}

\author{
\IEEEauthorblockN{
Di Dai$^{*}$,
Bo Liu$^{*}$,
Liwei Wang$^{\dagger}$
}

\IEEEauthorblockA{
\textit{State Key Laboratory of General Artificial Intelligence}\\
\textit{School of Intelligence Science and Technology}\\
\textit{Peking University}\\
Beijing, China\\
didai@stu.pku.edu.cn, liubo2022@stu.pku.edu.cn, wanglw@cis.pku.edu.cn
}

\IEEEauthorblockA{
$^{*}$Equal contribution.
\quad
$^{\dagger}$Corresponding author.
}
}

\maketitle

\begin{abstract}
Accurate lesion segmentation in mammography is important for computer-aided breast cancer diagnosis, but dense pixel-level annotations remain costly and limited. 
Existing public segmentation datasets usually cover only a small set of lesion types with manually delineated masks, whereas larger mammography datasets often provide only image-level diagnostic labels. 
This annotation mismatch makes it difficult to train segmentation models that can generalize to rare or weakly annotated lesion categories. 
To address this problem, we study open-vocabulary mammographic lesion segmentation under a mixed-supervision setting, where pixel-level masks are available for base lesion categories and image-level labels are available for a broader set of categories. 
We propose a two-branch framework that jointly learns from segmentation data and classification data. 
The segmentation branch learns lesion boundaries from annotated masks, while the classification branch introduces additional image-level semantic supervision. 
To better transfer image-level supervision to lesion-relevant regions, we further introduce a Lesion-wise Attentive Classifier that maintains category-specific feature prototypes and generates lesion-aware attention maps for foreground feature pooling. 
This design reduces the influence of background tissue and encourages more compact lesion representations. 
Experiments on multiple mammography datasets show that the proposed method consistently improves open-vocabulary segmentation baselines on both base and novel lesion categories, while reducing the gap to fully supervised models that require dense target-domain masks. 
Qualitative results further indicate that our method produces more complete lesion localization for small, low-contrast, and visually subtle mammographic findings. 
\end{abstract}


\section{Introduction}
Breast cancer is one of the most prevalent cancers among women and a leading cause of cancer-related mortality worldwide~\cite{dar2022breast,mambou2018breast,petrini2022breast,gotzsche2013screening}. Mammography is the standard imaging modality for breast cancer screening, and reliable interpretation of mammograms is crucial for early detection~\cite{blanks2020review}. A fundamental component of computer-aided diagnosis is precise lesion segmentation, which provides clear boundaries for masses, microcalcifications, and architectural distortions~\cite{azam2021mammographic}. Accurate segmentation provides reliable delineation of lesions, which supports quantitative analysis and clinical assessment~\cite{strudel2021segmenter,Kirillov_2023_ICCV}. Fully supervised methods trained with pixel-level annotations have shown strong performance~\cite{yuan2020object,chen2017deeplab,badrinarayanan2017segnet,long2015fully}. However, obtaining accurate lesion masks requires substantial effort from expert radiologists. Each lesion mask may require several minutes of careful annotation by experienced radiologists. Public datasets such as CBIS-DDSM~\cite{lee2017curated} and INBreast~\cite{moreira2012inbreast} provide only a limited number of annotated lesions. Models trained on these datasets often struggle to generalize to unseen datasets or rare lesion types. This scarcity of pixel-level labels remains a significant bottleneck in developing robust and scalable computer-aided diagnosis systems.

Recent studies have explored the use of image-level labels, which indicate the presence of lesions without precise pixel-level boundaries. Image-level labels are available from radiology reports or large screening datasets, which provide more diverse lesion examples for training. Weak supervision from image-level labels often produces coarse or incomplete localization, particularly for small lesions, microcalcifications, or overlapping structures. Lesion pixels may be misclassified as background without detailed guidance, which reduces the diagnosis efficiency. Integrating pixel-level and image-level labels allows the model to learn more accurate and generalizable lesion representations.

In this work, we propose a framework that jointly leverages pixel-level masks and image-level labels in a unified two-branch architecture. The segmentation branch is trained on mask annotations to learn precise lesion boundaries, while the classification branch exploits image-level supervision to enrich feature representations and provide coarse localization signals. A lesion-type-aware attention mechanism is used to focus on pixels likely belonging to each lesion type and suppress background regions. Focusing on lesion-relevant regions allows the model to generate more accurate segmentations, including for lesion types with only image-level labels. We categorize lesions with pixel-level masks as base categories, and lesions appearing only in image-level datasets as novel categories. This distinction helps the model integrate information from fine-grained annotations and broader image-level supervision. Joint training encourages feature clustering according to lesion type, which improves segmentation accuracy. The lesion-type-aware attention mechanism enhances feature alignment for each lesion type, which makes the model more sensitive to subtle lesions frequently missed by conventional methods. We conduct extensive evaluations on multiple mammography datasets. The segmentation branch is tested on CBIS-DDSM~\cite{lee2017curated} and INBreast~\cite{moreira2012inbreast}, while the image-level supervision leverages larger datasets CMMD~\cite{cui2021chinese} and VinDr-Mammo~\cite{nguyen2023vindr}. Experimental results show that our approach narrows the gap to fully supervised models while requiring fewer dense pixel-level annotations, which achieves higher segmentation results across base and novel categories. The model demonstrates strong generalization to unseen datasets and rare lesion types, which highlights the advantage of integrating heterogeneous supervision with lesion-type-aware attention. The code is avaliable at https://github.com/Ericdd90/LA-Classifier.

Our main contributions are summarized as follows:
\begin{itemize}
    \item We formulate mammographic lesion segmentation as a mixed-supervision open-vocabulary problem, where dense masks are available only for base lesion categories and image-level labels provide additional supervision for novel categories.
    \item We propose a unified two-branch framework that jointly exploits pixel-level segmentation data and image-level classification data, which allows the segmentation model to benefit from larger mammography datasets without requiring dense annotations.
    \item We introduce a Lesion-wise Attentive Classifier that uses category-specific memory features to highlight lesion-relevant foreground regions and suppress background tissue during image-level supervision.
    \item Experiments on multiple mammography datasets demonstrate the effectiveness on both base and novel lesion categories.
\end{itemize}

\section{Related Work}
\label{sec:related_works}

\subsection{Breast Cancer Classification}
Deep learning techniques have demonstrated significant advancements in breast cancer classification. Convolutional neural networks and transformer-based architectures~\cite{dosovitskiy2020image} have been widely employed to predict whether a mammogram contains malignant or benign lesions. These models leverage hierarchical feature representations to capture local and global patterns in breast tissue, which enables robust discrimination between normal and abnormal regions. The large-scale datasets like CMMD~\cite{cui2021chinese} and VinDr-Mammo~\cite{nguyen2023vindr} has facilitated the training of high-performance classification models. Image-level labels from these datasets allow models to learn discriminative features across diverse patient populations, breast densities, and imaging devices. Recent studies report classification accuracies exceeding 90\% on benchmark datasets~\cite{murty2024integrative}, which highlights the potential of deep learning as a supportive tool in mammography-based diagnosis. Despite these successes, image-level classification models inherently lack precise localization of lesions. This limitation restricts their integration into workflows that require region-specific assessments~\cite{hu2021region}.

\subsection{Lesion Segmentation}

Lesion segmentation aims to delineate the precise boundaries of abnormalities in medical images and plays a crucial role in computer-aided diagnosis systems. Accurate lesion masks provide fine-grained structural information that supports clinical tasks such as tumor characterization, treatment planning, and longitudinal disease monitoring. 
In mammography analysis, segmentation helps reveal the shape, margin, and spatial extent of lesions, which are important visual cues for radiologists when differentiating benign from malignant findings.

Fully supervised segmentation models based on convolutional neural networks have achieved promising performance on mammography datasets such as CBIS-DDSM~\cite{lee2017curated} and INBreast~\cite{moreira2012inbreast}. 
Architectures including U-Net~\cite{ronneberger2015u}, DeepLab~\cite{chen2017deeplab}, and SegNet~\cite{badrinarayanan2017segnet} have become widely adopted baselines due to their ability to capture multi-scale contextual information while preserving spatial resolution. 
Several studies further improved segmentation performance through architectural refinements such as attention mechanisms~\cite{wang2020self}, feature pyramid structures~\cite{varlamova2024features}, and encoder–decoder variants specifically designed for medical imaging tasks~\cite{bai2025qwen25vltechnicalreport}. 
These methods enable accurate localization of lesions and provide valuable information for downstream diagnostic models.

Despite these advances, mammographic lesion segmentation remains challenging due to the complex visual characteristics of breast X-ray images. Lesions often appear with low contrast relative to surrounding tissue, and their appearance varies significantly across patients and imaging conditions. 
For example, masses may exhibit irregular shapes and fuzzy margins, while microcalcifications typically appear as extremely small and sparse bright spots. Dense breast tissue can obscure lesion boundaries, further complicating the segmentation task. 
These factors make it difficult for models to generalize across datasets and clinical environments~\cite{shen2019deep,he2020deep}. Another critical limitation of fully supervised segmentation methods is their reliance on pixel-level annotations. 
Creating accurate lesion masks requires experienced radiologists to manually outline abnormal regions on high-resolution mammograms, which is both time-consuming and labor-intensive. 
As a result, the scale of publicly available segmentation datasets remains relatively limited compared with classification datasets that only require image-level labels. 
This annotation bottleneck restricts the development and generalization capability of supervised segmentation models, especially when dealing with rare lesion types or subtle abnormalities~\cite{dembrower2020multi,liberman2002breast}.

\subsection{Multi-Task Learning}
Multi-task learning has been widely explored in computer vision to jointly optimize multiple related tasks within a unified framework~\cite{caruana1997multitask,ruder2017overview,zhao2023multi}. 
By sharing feature representations across tasks, multi-task models can improve representation learning and enhance generalization performance. 
Recent deep learning studies further demonstrate the effectiveness of multi-task learning for various visual recognition problems~\cite{kendall2018multi,misra2016cross}. 
Several modern vision architectures adopt multi-task learning to simultaneously perform tasks such as classification, detection, and segmentation within a shared backbone network~\cite{he2017mask}. 
Transformer-based architectures have further advanced multi-task representation learning by enabling flexible parameter sharing and task-specific adaptation~\cite{tian2023end,zhong2024transforming}. 
These approaches show that jointly optimizing multiple tasks can improve feature learning and model robustness across diverse visual tasks. 
However, effectively balancing task-specific objectives and utilizing heterogeneous annotations from different datasets remain challenging problems.

\begin{figure*}[!t]
    \centering
    \includegraphics[width=\linewidth]{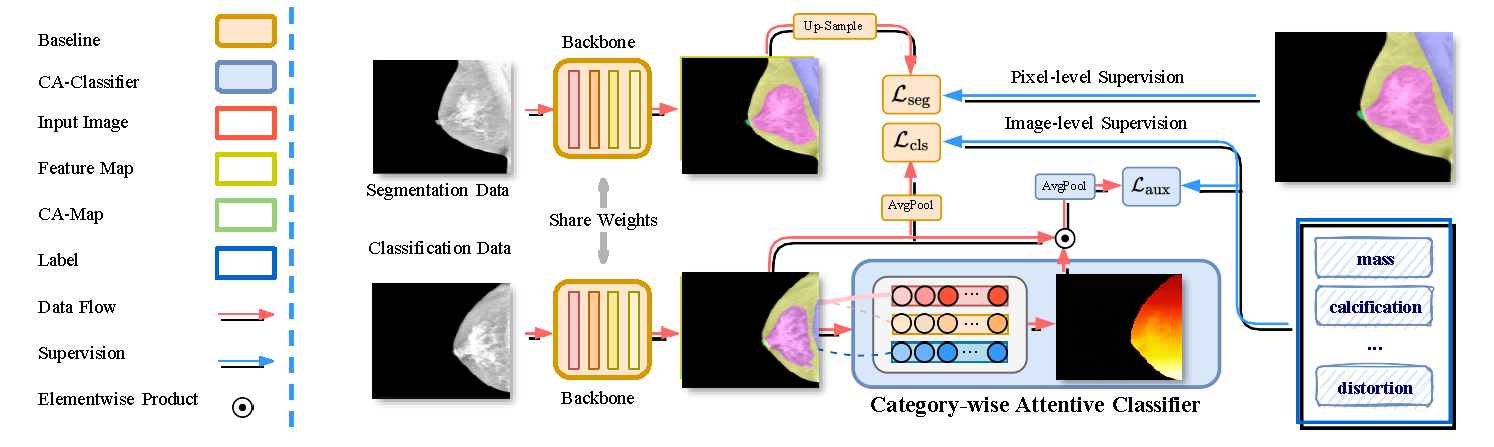}
    \caption{Overview of the proposed framework for mammography images. 
    The simple baseline jointly learns from segmentation and image-level classification data via pixel-level and image-level objectives ($\mathcal{L}_{\text{seg}}$, $\mathcal{L}_{\text{cls}}$). 
    The proposed Lesion-wise Attentive Classifier (LA-Classifier) maintains lesion-wise features in a memory bank to highlight foreground lesion pixels and suppress background. 
    The attentively pooled score map is supervised by an auxiliary image-level classification loss ($\mathcal{L}_{\text{aux}}$). LA-Map denotes the lesion-wise attention map.}
    \label{fig:framework}
\end{figure*}

\section{Problem Formulation}

We study mammographic lesion segmentation under a mixed-supervision open-vocabulary setting. 
Let $\mathcal{C}=\mathcal{C}_{base}\cup\mathcal{C}_{novel}$ denote the complete lesion vocabulary, where $\mathcal{C}_{base}$ contains lesion categories with pixel-level mask annotations during training, and $\mathcal{C}_{novel}$ contains categories for which only image-level labels are available during training. 
The goal is to learn a segmentation model that can produce pixel-wise predictions for both base and novel lesion categories.

We define the segmentation dataset as
\begin{equation}
\mathcal{D}_{seg}=\{(x_i, m_i, y_i)\}_{i=1}^{N_s},
\end{equation}
where $x_i\in\mathbb{R}^{H\times W}$ is a mammogram, $m_i\in\{0,1\}^{H\times W\times |\mathcal{C}_{base}|}$ is the pixel-level lesion mask, and $y_i$ is the corresponding image-level lesion label. 
For categories in $\mathcal{C}_{base}$, dense masks are used to supervise the segmentation branch.

We also use image-level classification datasets
\begin{equation}
\mathcal{D}_{cls}=\{(x_j, y_j)\}_{j=1}^{N_c},
\end{equation}
where $x_j$ is a mammogram and $y_j\in\{0,1\}^{|\mathcal{C}|}$ denotes the image-level label. 
Unlike $\mathcal{D}_{seg}$, these datasets do not provide pixel-level lesion masks. 
They are therefore used to provide weak semantic supervision for lesion categories that are under-represented or absent in the segmentation dataset.

During training, masks of novel categories are not used as pixel-level supervision. 
During evaluation, ground-truth masks of novel categories are used only for measuring segmentation performance. 
This setting reflects a practical scenario in mammography, where dense lesion contours are expensive to obtain, but image-level diagnostic labels are more widely available.

\section{Methodology}
\label{sec:method}

\subsection{Framework Overview}

The proposed framework consists of a shared visual encoder, a segmentation branch, and a lesion-wise classification branch, as shown in Figure~\ref{fig:framework}. 
Given an input mammogram $x$, the shared encoder $E_{\theta}$ extracts a feature map
\begin{equation}
F=E_{\theta}(x), \quad F\in\mathbb{R}^{h\times w\times d},
\end{equation}
where $h$ and $w$ denote the spatial resolution of the feature map and $d$ is the feature dimension.

The segmentation branch predicts pixel-wise lesion probabilities from $F$. 
It is supervised by pixel-level masks from $\mathcal{D}_{seg}$ and is responsible for learning lesion boundaries and spatial structures. 
The classification branch predicts image-level lesion labels from pooled features and is supervised by $\mathcal{D}_{cls}$. 
Because image-level labels do not indicate where lesions appear, directly applying global average pooling may assign supervision to both lesion and background regions. 
This issue is particularly severe in mammography, where dense tissue, glandular structures, and imaging artifacts may co-occur with suspicious findings.

To address this limitation, the Lesion-wise Attentive Classifier estimates category-specific attention maps from feature similarities. 
These attention maps are used to enhance lesion-relevant foreground regions before image-level classification. 
As a result, image-level supervision is encouraged to focus on pixels that are visually consistent with each lesion category, rather than being uniformly distributed over the entire mammogram. 
The framework is trained end-to-end with segmentation, classification, and auxiliary attention-guided classification losses.

\subsection{Baseline}
\label{subsec:baseline}

\noindent\textbf{Incorporating Image-Level Supervision.}
Our baseline segmenter consists of a backbone network (e.g., ViT-B/16~\cite{dosovitskiy2020image}) followed by a single-layer cosine classifier for pixel-wise category prediction. 
To incorporate image-level supervision, an additional image classification branch is introduced by applying global average pooling to the feature map, followed by a classification head. 
For images containing multiple categories, the cross-entropy loss is computed independently for each present category. 
To avoid conflicts among categories, all other categories are treated as ignore labels and are excluded from the loss computation.

\noindent\textbf{Joint Training.}
To enable the segmentation model to recognize \texttt{novel} categories, the network is jointly trained using both segmentation and classification data. 
The backbone parameters are shared across the two tasks. 
The overall training objective is defined as
\begin{align}
\mathcal{L}=\mathcal{L}_{\text{seg}}+\lambda_{\text{cls}} \mathcal{L}_{\text{cls}},
\end{align}
where $\lambda_{\text{cls}}$ is a weighting coefficient controlling the contribution of the classification loss, which is set to $0.1$ by default.

\noindent\textbf{Inference.}
During inference, pixel-wise category predictions are obtained directly from the segmentation head. 
Since the backbone parameters are jointly optimized with classification supervision during training, the model can naturally segment \texttt{novel} categories that only appear in the classification dataset.

\subsection{Analysis of Novel Category Pixel Grouping}
\label{subsec:subtask_analysis}

\begin{table}[!t]
    \setlength{\tabcolsep}{3pt}
    \centering
    \caption{Results of the exploratory experiment. For the second row, the evaluation categories are not in the training vocabulary. For the third row, the evaluation categories are within the training vocabulary. CLIP-RN50x64 with VILD prompt engineering is used to generate DenseCLIP~\cite{zhou2021denseclip} results.}
    \resizebox{\linewidth}{!}
{
    \begin{tabular}{l|lc|lc}
        \toprule
        Model & Training Data & Test Data & mIoU \\
        \midrule
        \midrule
        DenseCLIP & ITPair & Mammo-bench$-$CBIS-DDSM & 13.6 \\
        Basic Segmenter & CBIS-DDSM & Mammo-bench$-$CBIS-DDSM & 73.6 \\
        Basic Segmenter & Mammo-bench & Mammo-bench$-$CBIS-DDSM & \color{gray}{76.2} \\
        \midrule
        DenseCLIP & ITPair & Mammo-bench$-$INBreast & 10.8 \\
        Basic Segmenter & INBreast & Mammo-bench$-$INBreast & 68.6 & \\
        Basic Segmenter & Mammo-bench & Mammo-bench$-$INBreast & \color{gray}{73.4} \\
        \bottomrule
    \end{tabular}}
    \label{tab:oracle_exp}
\end{table}

Despite the strong performance of the proposed baseline, it still suffers from a potential limitation. 
The probability vector $\bar{\mathcal{P}}_{\text{cls}}$ is obtained via global average pooling, which implies that the classification supervision is applied only at the image level and cannot accurately localize the corresponding pixels. 
To address this issue, we aim to extract foreground regions of \texttt{novel} categories and apply fine-grained supervision to them. 
However, precise region extraction without pixel-level annotations requires strong intra-category pixel compactness. 
Therefore, we train a segmentation model on \texttt{base} categories and investigate whether such compactness also holds for \texttt{novel} categories.

\noindent\textbf{Exploratory Experiment.}
This experiment aims to measure intra-category pixel compactness in a category-agnostic manner. 
Specifically, a basic segmenter is trained using only \texttt{base} categories. 
During inference, for images containing \texttt{novel} category regions, we extract feature maps $\mathcal{F}$ and randomly select a pixel $(h_c,w_c)$ from the ground-truth mask of each category. 
The feature of the selected pixel is treated as the representative feature of that category. 
We then compute the cosine similarity between the representative feature $\mathcal{F}[h_c,w_c]$ and all other pixels in the feature map to obtain the pixel-level response map $\mathcal{S}_{\text{res}}$:
\begin{align}
\label{eq:response_map}
\mathcal{S}_{\text{res}}=\operatorname{cos}(\mathcal{F}[h_c,w_c], \mathcal{F}).
\end{align}

Higher response scores indicate stronger similarity to the representative feature. 
Each pixel is assigned to the category with the highest response score. Quantitative results are reported in Table~\ref{tab:oracle_exp}. 
Interestingly, categories outside the training vocabulary exhibit intra-category compactness comparable to those within the training vocabulary.

\noindent\textbf{Discussion.}
The results suggest that pixel grouping ability generalizes to out-of-distribution categories in a category-agnostic manner. 
This observation likely stems from the fact that pixels belonging to the same semantic category often share similar spatial patterns, colors, and textures. 
Consequently, pixel features learned from image classification data tend to form implicit semantic clusters. 
This property indicates that fine-grained supervision signals can potentially be derived from image-level labels, provided that pixel groups corresponding to different categories can be explicitly distinguished.

\subsection{Lesion-wise Attentive Classifier (LA-Classifier)}\label{subsec:background}

Based on the observation that pixel features of \texttt{novel} lesion categories naturally form compact clusters, we propose a \textbf{Lesion-wise Attentive Classifier (LA-Classifier)} to explicitly distinguish different pixel groups and extract foreground regions for fine-grained supervision.

\noindent\textbf{Memory-based Category Representation.} 
To capture cross-image semantic cues, we maintain a memory bank $\mathbb{M} = \{\mathcal{M}_i \mid i \in \text{\texttt{novel} lesions}\}$ storing representative features for each novel category. At each training step, we update the memory bank using pixel features with top-$K$ classification scores. The confidence of a pixel $(h,w)$ belonging to category $i$ is computed as the average cosine similarity between its feature $\mathcal{F}_{(h,w)}$ and the category's memory entries:
\begin{align}
    \mathcal{S}^{i}_{\text{mem}} = \operatorname{Avg}\left(\cos(\mathcal{F}_{(h,w)}, \mathcal{M}_{i})\right).
\end{align}

\noindent\textbf{Lesion-wise Attention Map.} 
Foreground strengthening and background suppression are applied to produce a lesion-wise attention map:
\begin{align}
    \mathcal{A} = \operatorname{sigmoid}(\operatorname{norm}(\mathcal{S}_{\text{mem}}^{\text{fg}} - \mathcal{S}_{\text{mem}}^{\text{bg}})),
\end{align}
where $\text{fg}$ and $\text{bg}$ denote foreground and background pixels. The attention map is then used to pool the feature map and generate an auxiliary image-level classification prediction:
\begin{align}
    \mathcal{P}^{i}=\operatorname{softmax}(\mathcal{S}^{i}_{\text{cls}} \cdot \mathcal{A} / \tau), \quad \forall i \in \text{\texttt{novel} lesions}.
\end{align}
In this way, image-level supervision is accurately assigned to the foreground pixel group, as illustrated in Fig.~\ref{fig:framework}.

\subsection{Training Objective}

For images from the segmentation dataset, the segmentation branch is optimized using a combination of cross-entropy loss and Dice loss:
\begin{equation}
\mathcal{L}_{seg} = \mathcal{L}_{ce}(\hat{m}, m) + \lambda_{dice}\mathcal{L}_{dice}(\hat{m}, m),
\end{equation}
where $\hat{m}$ and $m$ denote the predicted and ground-truth lesion masks, respectively. 
The cross-entropy term encourages pixel-wise category correctness, while the Dice term improves region-level overlap, which is important for small lesions and sparse calcification clusters.

For image-level classification data, the classification branch is trained with an image-level classification loss:
\begin{equation}
\mathcal{L}_{cls} = - \frac{1}{|\mathcal{Y}|}\sum_{c\in\mathcal{Y}}\log p_c,
\end{equation}
where $\mathcal{Y}$ denotes the set of positive image-level labels and $p_c$ is the predicted probability for category $c$. 
When only positive labels are provided, categories not annotated in the image are ignored rather than treated as negative labels. 
This avoids penalizing the model for unlabeled co-existing findings.

The auxiliary loss is applied to the attention-pooled classification score generated by the LA-Classifier:
\begin{equation}
\mathcal{L}_{aux} = - \frac{1}{|\mathcal{Y}|}\sum_{c\in\mathcal{Y}}\log p^{aux}_c,
\end{equation}
where $p^{aux}_c$ is obtained from the lesion-wise attentive feature pooling. 
The final objective is
\begin{equation}
\mathcal{L} = \mathcal{L}_{seg} + \lambda_{cls}\mathcal{L}_{cls} + \lambda_{aux}\mathcal{L}_{aux}.
\end{equation}
Following our default setting, $\lambda_{cls}$ and $\lambda_{aux}$ are set to 0.1 unless otherwise specified.

\section{Experiments}
\label{sec:experiments}

\subsection{Datasets}
\label{subsec:datasets}

We evaluate the proposed method on several publicly available mammography datasets that provide lesion annotations for breast cancer analysis. The summary
of all datasets is shown in Table~\ref{tab:datasets}.

\noindent\textbf{MIAS~\cite{suckling1994mammographic}.} The MIAS dataset is the earliest publicly available digitized screening mammography datasets that provided precise annotations of breast abnormalities (i.e., screen-film mammography).

\noindent\textbf{CBIS-DDSM~\cite{lee2017curated}.}
The Curated Breast Imaging Subset of DDSM is a large mammography dataset derived from the Digital Database for Screening Mammography.
It contains 3,000+ mammogram images with expert annotations for masses and calcifications. 
Pixel-level lesion masks are provided for segmentation tasks.

\noindent\textbf{INBreast~\cite{moreira2012inbreast}.}
The INBreast dataset is a high-quality full-field digital mammography dataset containing 410 images from 115 patients. 
The dataset provides precise lesion annotations including masses, calcifications, and asymmetries. 
Following common practice, we use the official lesion masks as ground-truth segmentation labels.

\noindent\textbf{CMMD~\cite{cui2021chinese}.} The Chinese Mammography Database is a large-scale clinical dataset containing 5,202 mammography images from 1,775 patients. Unlike the aforementioned datasets that provide pixel-level or bounding-box annotations, CMMD primarily offers image-level labels, including biopsy-confirmed pathology (benign or malignant) and breast density. It is widely used for evaluating the classification performance and generalizability of models on the Chinese population.

\noindent\textbf{VinDr-Mammo~\cite{nguyen2023vindr}.}
VinDr-Mammo is a recent large-scale mammography dataset including 5,000 exams with breast-level and lesion-level annotations.
Bounding boxes and lesion categories are provided by expert radiologists. 
For segmentation experiments, lesion regions are converted into pixel-level masks.

\noindent\textbf{Mammo-Bench~\cite{bhole2025mammo}.} The Mammo-Bench dataset is introduced as a large-scale, standardized benchmark for mammography. It contains 19,731 images, which provides comprehensive annotations including BI-RADS assessments, breast density, abnormality types, and ROI masks. 

\begin{table}[!t]
\centering
\caption{Existing Public Mammography Datasets}
\label{tab:datasets}
\resizebox{\linewidth}{!}
{
\begin{tabular}{ccccc}
\toprule
\textbf{Dataset} & \textbf{Year} & \textbf{Images} & \textbf{Annotation} & \textbf{BI-RADS} \\
\hline
MIAS \cite{suckling1994mammographic} & 1994 & 322 & Circle around the
finding & No \\
CBIS-DDSM \cite{lee2017curated} & 2017 & 3,103 & Contour enclosing the
findings & Yes \\
INBreast \cite{moreira2012inbreast} & 2012 & 410 & Contour enclosing the
finding & Yes \\
CMMD \cite{cui2021chinese} & 2021 & 5,202 & Pathological labels & No \\
VinDr-Mammo \cite{nguyen2023vindr} & 2022 & 20,000 & Rectangular region of
interest & Yes \\
Mammo-Bench \cite{bhole2025mammo} & 2025 & 19,731 & Region of interest mask
 & Yes \\
\bottomrule
\end{tabular}
}
\end{table}

\subsection{Experimental Setups}
\label{subsec:exp_settings}

\noindent\textbf{CMMD As Classification Data} 
We extend the vocabulary of CBIS-DDSM by incorporating the classification dataset CMMD and evaluate the resulting models on the INBreast dataset. 
CMMD provides large-scale mammography images with image-level diagnostic labels, which introduce additional lesion categories and appearance variations that are not fully covered by CBIS-DDSM. 
By integrating this dataset, the model is exposed to a broader range of lesion patterns and imaging conditions during training, enabling it to learn more diverse visual representations. 
Some lesion types appear much more frequently than others, which may bias the learning process and affect the model’s ability to generalize to rare categories. 
This imbalance makes the setting more challenging and highlights the importance of effectively utilizing weak supervision signals from classification data.

\noindent\textbf{Vindr-Mammo As Classification Data} 
We further incorporate the VinDr-Mammo dataset into the above training setting and evaluate the models on the INBreast dataset. 
VinDr-Mammo is a large-scale mammography dataset collected from multiple clinical centers and contains a diverse set of imaging conditions and patient populations. 
Compared with CMMD, VinDr-Mammo introduces additional variability in breast density, lesion morphology, and imaging quality, which can potentially improve the robustness of the learned representations. 
By combining CBIS-DDSM, CMMD, and VinDr-Mammo, the training process benefits from both pixel-level segmentation annotations and large-scale image-level labels. 
Evaluating the resulting models on INBreast provides a challenging cross-dataset validation scenario that reflects the generalization capability of the proposed approach.
 
\noindent\textbf{Mammo-Bench as External Evaluation.}
To further evaluate cross-dataset generalization, we use Mammo-Bench as an external evaluation dataset, which is not used during training in this setting. Mammo-Bench contains mammography images with annotated abnormalities and provides additional lesion examples that are not fully represented in CBIS-DDSM. 
By integrating this dataset, the model can observe a wider range of lesion appearances and anatomical variations, which helps improve its ability to recognize diverse abnormalities.  
This experiment also provides insights into how expanding the segmentation vocabulary affects the model’s performance on unseen datasets and lesion categories.

\subsection{Evaluation Protocol}

We evaluate segmentation performance using mean Intersection-over-Union (mIoU). 
For each lesion category $c$, IoU is computed as
\begin{equation}
\mathrm{IoU}_c = \frac{TP_c}{TP_c+FP_c+FN_c},
\end{equation}
where $TP_c$, $FP_c$, and $FN_c$ denote the number of true-positive, false-positive, and false-negative pixels for category $c$, respectively. 
The final mIoU is obtained by averaging over evaluated categories.

We report three groups of results. 
``All'' denotes the average performance over all evaluated lesion categories. 
``Base'' denotes categories with pixel-level mask supervision during training. 
``Novel'' denotes categories that are not provided with pixel-level masks during training and are introduced through image-level labels or held-out evaluation categories. 
This evaluation protocol allows us to separately examine standard segmentation performance and generalization to weakly supervised lesion categories.

All images are resized following the implementation settings described below. 
Unless otherwise stated, results are reported without multi-scale testing, dense CRF post-processing, or test-time augmentation. 
When patient identifiers are available, dataset splits are performed at the patient level to avoid leakage between training and testing. 
For datasets with bounding-box annotations, box-derived regions are used only according to the specified supervision setting, and evaluation is performed on the available lesion masks or ROI annotations.

\subsection{Implementation Details}
\label{subsec:impl_details}

\noindent\textbf{Overall Framework.} 
We use ViT-B/16~\cite{dosovitskiy2020image} initialized with I21K pre-trained weights as the backbone. All models are trained on $8$ V100 GPUs. Following Detic~\cite{zhou2022detecting}, images from the same dataset are grouped on the same GPU to improve training efficiency. For all experiments, two images are sampled per mini-batch per GPU. For setups (1) and (3), we use $4 : 4$ GPUs for
segmentation and classification data. For setup (2), we use $3 : 3 : 2$ GPUs for
Vindr-Mammo, CBIS-DDSM and INBreast.

\noindent\textbf{Training Parameters.} 
Models for the main results and ablation study are trained for $320$K iterations. For large-vocabulary segmentation using INBreast, we train for $1280$K iterations. 
Images from CBIS-DDSM, CMMD, and VinDr-Mammo datasets are resized to $512 \times 512$, while images from MIAS and INBreast are resized to $320 \times 320$. 
Data augmentations include multi-scale jittering with a random scale in $[0.5, 2.0]$, random cropping, random flipping, and photometric distortion. 
We adopt SGD with a base learning rate of $0.001$, momentum $0.9$, and no weight decay. 
The learning rate follows a polynomial schedule with power $0.9$ and minimum $1\times10^{-5}$. 
We empirically set the loss weights $\lambda_{\text{cls}}=0.1$ and $\lambda_{\text{aux}}=0.1$. 
All results are evaluated without multi-scale testing.

\subsection{Main Results}
\label{subsec:results}

\begin{table*}[!t]
\centering
\caption{\textbf{Comparison of baselines and our proposed models and supervised learning models on mammography datasets.} The supervised learning results are shown in \color{gray}{gray}\color{black}. CBIS-DDSM and INBreast are used as segmentation data; CMMD and VinDr-Mammo are used as classification data. ITPair stands for the web-scale image-text pairs to train a CLIP/ALIGN model. All models are evaluated without multi-scale testing and Dense CRF post-processing.}
\resizebox{\linewidth}{!}{
\begin{tabular}{ll|lccc|lccc}
\toprule
\multirow{2}{*}{Model} & \multirow{2}{*}{Backbone} & \multirow{2}{*}{Training Data} & \multicolumn{3}{c|}{MIAS-mIoU} & \multirow{2}{*}{Training Data} & \multicolumn{3}{c}{INBreast-mIoU} \\
& & & All & Base & Novel & & All & Base & Novel\\
\midrule
\midrule
DenseCLIP~\cite{zhou2021denseclip} & RN50x64~\cite{he2016deep} & ITPairs & 12.4 & 16.3 & 10.2 & ITPairs & 5.3 & 9.6 & 4.8 \\
ReCo~\cite{shin2022reco} & RN50x64 & ITPairs & 13.8 & 16.7 & 11.4 & ITPairs & 5.9 & 9.8 & 5.1 \\

ZSBaseline~\cite{xu2021simple} & ViT-B/16 & CBIS-DDSM & 61.9 & 66.4 & 58.7 & CBIS-DDSM & 60.5 & 65.1 & 57.9 \\
LSeg~\cite{li2022language} & ViT-B/16 & CBIS-DDSM & 63.2 & 68.1 & 59.6 & CBIS-DDSM & 61.7 & 66.3 & 58.8 \\
OpenSeg~\cite{ghiasi2021open} & EfficientNet-B7 & CBIS-DDSM & 65.4 & - & - & CBIS-DDSM & 63.8 & - & - \\

\midrule
Baseline & ViT-B/16 & CBIS-DDSM, CMMD & 66.8 & 73.6 & 61.2 & CBIS-DDSM, CMMD & 63.4 & 72.1 & 59.7\\
Ours & ViT-B/16 & CBIS-DDSM, CMMD & \textbf{70.2} & \textbf{75.4} & \textbf{65.5} & CBIS-DDSM, CMMD & \textbf{66.7} & \textbf{73.5} & \textbf{62.8}\\

\midrule
Baseline & ViT-B/16 & CBIS-DDSM, CMMD, VinDr-Mammo & 69.9 & 74.3 & 64.7 & CBIS-DDSM, CMMD, VinDr-Mammo & 67.5 & 73.8 & 63.9\\
Ours & ViT-B/16 & CBIS-DDSM, CMMD, VinDr-Mammo & \textbf{72.8} & \textbf{76.2} & \textbf{68.1} & CBIS-DDSM, CMMD, VinDr-Mammo & \textbf{70.1} & \textbf{75.6} & \textbf{66.8}\\

\midrule
Supervised & ViT-B/16 & INBreast & \color{gray}{75.6} & \color{gray}{80.2} & \color{gray}{70.4} & INBreast & \color{gray}{71.8} & \color{gray}{78.7} & \color{gray}{68.9}\\
\bottomrule
\end{tabular}
}
\label{tab:sota_comp_mammo_full}
\end{table*}

\noindent\textbf{CMMD As Classification Data.} 
In Table~\ref{tab:sota_comp_mammo_full}, we compare our method with existing language-guided segmentation approaches on mammography datasets. 
When extending the vocabulary of CBIS-DDSM with the classification dataset CMMD, the simple baseline already outperforms previous open-vocabulary methods such as DenseCLIP~\cite{zhou2021denseclip}, ReCo~\cite{shin2022reco}, LSeg~\cite{li2022language}, and OpenSeg~\cite{ghiasi2021open} by a large margin. 
This demonstrates that incorporating additional mammography classification data effectively enlarges the visual vocabulary for open-vocabulary segmentation. 
Furthermore, our method consistently improves over the baseline model, bringing notable gains on both MIAS and INBreast benchmarks, especially on \texttt{novel} categories. 
These results indicate that the proposed design improves the generalization ability to unseen categories.

\noindent\textbf{CMMD and VinDr-Mammo As Classification Data.} 
We further include VinDr-Mammo as an additional classification dataset during training. 
Compared with using CMMD alone, adding VinDr-Mammo further improves the segmentation performance on both evaluation datasets. 
Specifically, the baseline model shows clear improvements on MIAS and INBreast, suggesting that more diverse mammography images provide richer visual supervision. 
With our proposed method, the performance is further improved, achieving the best results across all metrics. 
These results demonstrate that enlarging the training vocabulary with multiple mammography datasets is beneficial for open-vocabulary segmentation.

\noindent\textbf{Clarification of supervised references.}
The fully supervised results shown in gray are included as reference upper-bound results rather than direct open-vocabulary baselines. 
These models are trained with dense pixel-level masks from the corresponding segmentation datasets, and therefore assume a stronger annotation setting than the proposed mixed-supervision framework. 
In contrast, our method focuses on the setting where dense masks are available only for base categories, while additional lesion categories are introduced through image-level labels. 
Therefore, the main comparison should be made against open-vocabulary and mixed-supervision baselines under the same annotation setting. 
The supervised results indicate the performance gap that remains when dense target-domain masks are available for training.

\subsection{Ablation Study}
\label{subsec:ablation}

\noindent\textbf{Effectiveness of LA-Classifier.} 
We first analyze the effectiveness of the proposed LA-Classifier under different classification data settings. 
Specifically, we apply the LA-Classifier to CMMD and VinDr-Mammo classification datasets respectively. 
As shown in Table~\ref{tab:ca_cls_effective}, applying the LA-Classifier to either dataset improves the segmentation performance compared with the baseline. 
The best performance is achieved when the LA-Classifier is applied to both datasets simultaneously. 
This result indicates that leveraging lesion-aware supervision from multiple classification datasets is beneficial for improving open-vocabulary mammography segmentation.

\begin{table}[!t]
\centering
\caption{Ablation study on applying the LA-Classifier to different classification datasets. LAC-CMMD and LAC-VinDr-Mammo denote using LA-Classifier on CMMD and VinDr-Mammo respectively.}
{
\begin{tabular}{cccccc}
\toprule
LAC-CMMD & LAC-VinDr-Mammo & All & Base & Novel\\
\midrule
\midrule
- & - & 71.3 & 78.7 & 66.3 \\
- & $\checkmark$ & 72.4 & 79.1 & 67.8\\
$\checkmark$ & - & 73.6 & 80.2 & 69.5\\
$\checkmark$ & $\checkmark$ & \textbf{74.2} & \textbf{80.9} & \textbf{70.6}\\
\bottomrule
\end{tabular}
}
\label{tab:ca_cls_effective}
\end{table}

\noindent\textbf{Design of LA-Classifier.} 
We further compare our cross-image design with a simpler single-image design. 
In the single-image variant, high-score pixels from the classification score map are directly selected as foreground regions and supervised with cross-entropy loss. 
As shown in Table~\ref{tab:design}, the proposed cross-image LA-Classifier significantly outperforms the single-image design. 
This improvement mainly comes from the ability to leverage semantic information across different images, which helps suppress background regions that frequently co-occur with lesions in mammography images.

\begin{table}[!t]
\centering
\caption{Comparison of different LA-Classifier designs. Single-image selects foreground regions from classification score maps, while Cross-image leverages cross-image semantic information (our method).}
{
\begin{tabular}{lcccc}
\toprule
LA-Classifier Design & All & Base & Novel\\
\midrule
\midrule
Single-Image & 70.1 & 78.6 & 64.8 \\
Cross-Image & \textbf{72.4} & \textbf{81.3} & \textbf{67.8}\\
\bottomrule
\end{tabular}
}
\label{tab:design}
\end{table}

\noindent\textbf{Ablation on Memory Bank.} 
We also study the influence of the memory bank size $M$ and the top-$K$ region selection strategy. 
We first vary the memory bank size while fixing the top-$K$ region area to 40, as shown in the left part of Table~\ref{tab:memory_bank}. 
The model achieves the best performance when $M=20$. 
Next, we fix the memory bank size to $20$ and vary the top-$K$ region area. 
As shown in the right part of Table~\ref{tab:memory_bank}, the best performance is obtained when $K=20$. 
Therefore, we adopt $M=20$ and $K=20$ in all experiments.

\begin{table}[!t]
\centering
\caption{Ablation on memory bank size $M$ (left) and top-$K$ region selection (right). We select the hyper-parameters with the best performance on novel categories.}
{
\begin{tabular}{cccc|cccc}
\toprule
Size & All & Base & Novel & Top-K & All & Base & Novel\\
\midrule
\midrule
10 & \textbf{72.6} & \textbf{82.8} & 76.2 & 10 & 71.6 & 80.3 & 76.2\\
20 & 72.0 & 80.5 & \textbf{76.7} & 20 & \textbf{72.4} & 81.3 & \textbf{76.8}\\
40 & 72.2 & 82.0 & 76.1 & 40 & 72.2 & \textbf{82.0} & 76.1\\
80 & 71.6 & 80.9 & 75.7 & 80 & 71.4 & 80.0 & 75.9\\
\bottomrule
\end{tabular}
}
\label{tab:memory_bank}
\end{table}

\subsection{Error Analysis}
\label{sec:error_analysis}

To better understand the behavior of the proposed method, we conduct a detailed error analysis on the held-out segmentation masks.  The analysis is designed to answer two questions: (i) which types of errors are most effectively reduced by lesion-wise attentive supervision, and (ii) which lesion properties still make open-vocabulary mammographic segmentation difficult.  We decompose the prediction errors into four groups: false-negative errors, false-positive errors, boundary errors, and category-confusion errors.

Let $m_c \in \{0,1\}^{H\times W}$ and $\hat m_c \in \{0,1\}^{H\times W}$ denote the ground-truth mask and predicted mask for lesion category $c$, respectively.  False-negative and false-positive errors are defined as
\begin{align}
N_{\mathrm{FN}}^c &= \left|\{p: m_c(p)=1,\, \hat m_c(p)=0\}\right|, \\
N_{\mathrm{FP}}^c &= \left|\{p: m_c(p)=0,\, \hat m_c(p)=1\}\right| .
\end{align}
To measure contour quality, we count boundary mismatches within a narrow contour band.  Specifically, let $\mathcal{B}_{\rho}(m_c)$ denote a $\rho$-pixel band around the boundary of $m_c$.  The boundary error is computed as
\begin{equation}
N_{\mathrm{Bnd}}^c = \left| (m_c \oplus \hat m_c) \cap \mathcal{B}_{\rho}(m_c) \right|,
\end{equation}
where $\oplus$ denotes the pixel-wise exclusive-or operation.  For multi-category segmentation, we further count category-confusion errors as lesion pixels that are assigned to an incorrect lesion category:
\begin{equation}
\begin{aligned}
N_{\mathrm{Cls}}^c = \big|\{p: &\; m_c(p)=1,\, \exists c'\neq c,\\
&\; \hat m_{c'}(p)=1\}\big| .
\end{aligned}
\end{equation}
The normalized ratio of an error type $t$ is reported as
\begin{equation}
 r_t = \frac{\sum_{c} N_t^c}{\sum_{u\in\{\mathrm{FN},\mathrm{FP},\mathrm{Bnd},\mathrm{Cls}\}}\sum_{c} N_u^c}.
\end{equation}

Fig.~\ref{fig:error_analysis} summarizes the major error patterns.  Compared with the mixed-supervision baseline based on global image-level pooling, our method reduces missed lesion regions and suppresses background activations more effectively.  This trend is especially clear for subtle microcalcification clusters and low-contrast masses, where global image-level supervision often produces incomplete responses.  In dense breast tissue, the baseline frequently activates normal glandular structures as false positives, while the proposed lesion-wise attentive classifier generates more compact lesion responses by pooling foreground features around category-specific memory prototypes.

We also analyze failure distribution with respect to lesion size and local contrast.  For each connected lesion component, we measure the lesion area $s=|m_c|$ and the local contrast
\begin{equation}
\kappa = \frac{|\mu_{\mathrm{in}}-\mu_{\mathrm{out}}|}{\sigma_{\mathrm{out}}+\epsilon},
\end{equation}
where $\mu_{\mathrm{in}}$ is the mean intensity inside the lesion, and $\mu_{\mathrm{out}}$ and $\sigma_{\mathrm{out}}$ are computed from a surrounding local background ring.  Remaining errors are concentrated in the small-size and low-contrast region.  Extremely sparse calcifications may still be missed, and lesions with fuzzy margins may lead to partial boundary under-segmentation or over-segmentation.  These results indicate that future improvements may benefit from higher-resolution decoding, multi-view consistency, and uncertainty-aware boundary refinement.

\definecolor{figTextGray}{HTML}{555555}
\newcommand{\figTitle}{\fontsize{8.0}{8.8}\selectfont\bfseries}
\newcommand{\figAxis}{\fontsize{6.7}{7.3}\selectfont}
\newcommand{\figTick}{\fontsize{6.1}{6.7}\selectfont}
\newcommand{\figLegend}{\fontsize{6.2}{6.9}\selectfont}
\newcommand{\figNote}{\fontsize{6.0}{6.6}\selectfont}

Fig.~\ref{fig:error_analysis} summarizes the major error patterns through two complementary views. Panel~(a) decomposes pixel-level errors into false-negative, false-positive, boundary, and category-confusion errors. Compared with the mixed-supervision baseline based on global image-level pooling, our method reduces missed lesion regions and suppresses background activations more effectively. This trend is especially clear for subtle microcalcification clusters and low-contrast masses, where global image-level supervision often produces incomplete responses. In dense breast tissue, the baseline frequently activates normal glandular structures as false positives, whereas the proposed lesion-wise attentive classifier generates more compact lesion responses by pooling foreground features around category-specific memory prototypes.

We also analyze failure distribution with respect to lesion size and local contrast in Panel~(b). For each connected lesion component, we measure the lesion area $s=|m_c|$ and the local contrast
\begin{equation}
\kappa = \frac{|\mu_{\mathrm{in}}-\mu_{\mathrm{out}}|}{\sigma_{\mathrm{out}}+\epsilon},
\end{equation}
where $\mu_{\mathrm{in}}$ is the mean intensity inside the lesion, and $\mu_{\mathrm{out}}$ and $\sigma_{\mathrm{out}}$ are computed from a surrounding local background ring. Remaining errors are concentrated in the small-size and low-contrast region. Extremely sparse calcifications may still be missed, and lesions with fuzzy margins may lead to partial boundary under-segmentation or over-segmentation. These results indicate that future improvements may benefit from higher-resolution decoding, multi-view consistency, and uncertainty-aware boundary refinement.

\begin{figure*}[!t]
\centering
\begin{overpic}[width=\textwidth]{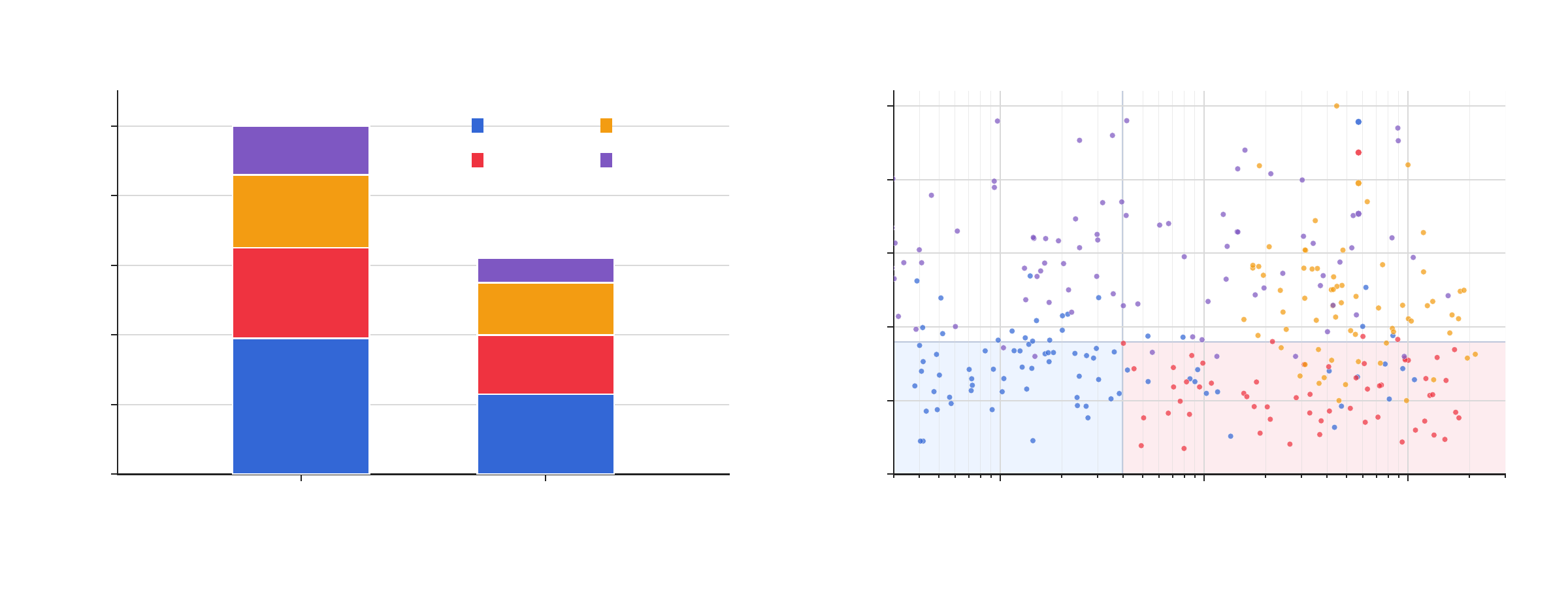}
  \put(27.00,35.49){\makebox(0,0)[c]{{\figTitle (a) Error type decomposition}}}
  \put(76.50,35.49){\makebox(0,0)[c]{{\figTitle (b) Failure distribution by lesion property}}}

  \put(2.70,20.73){\rotatebox{90}{\makebox(0,0)[c]{{\figAxis Normalized error pixels}}}}
  \put(18.88,6.76){\makebox(0,0)[c]{{\figTick Baseline}}}
  \put(35.13,6.76){\makebox(0,0)[c]{{\figTick Ours}}}

  \put(6.65,8.53){\makebox(0,0)[r]{{\figTick 0}}}
  \put(6.65,12.96){\makebox(0,0)[r]{{\figTick 20}}}
  \put(6.65,17.40){\makebox(0,0)[r]{{\figTick 40}}}
  \put(6.65,21.84){\makebox(0,0)[r]{{\figTick 60}}}
  \put(6.65,26.28){\makebox(0,0)[r]{{\figTick 80}}}
  \put(6.65,30.72){\makebox(0,0)[r]{{\figTick 100}}}

  \put(31.25,30.74){\makebox(0,0)[l]{{\figLegend False Neg.}}}
  \put(31.25,28.54){\makebox(0,0)[l]{{\figLegend False Pos.}}}
  \put(39.45,30.74){\makebox(0,0)[l]{{\figLegend Boundary}}}
  \put(39.45,28.54){\makebox(0,0)[l]{{\figLegend Confusion}}}

  \put(52.20,20.73){\rotatebox{90}{\makebox(0,0)[c]{{\figAxis Local contrast}}}}
  \put(76.50,4.86){\makebox(0,0)[c]{{\figAxis Lesion area (pixels, log scale)}}}

  \put(56.00,8.53){\makebox(0,0)[r]{{\figTick 0.0}}}
  \put(56.00,13.22){\makebox(0,0)[r]{{\figTick 0.1}}}
  \put(56.00,17.91){\makebox(0,0)[r]{{\figTick 0.2}}}
  \put(56.00,22.61){\makebox(0,0)[r]{{\figTick 0.3}}}
  \put(56.00,27.30){\makebox(0,0)[r]{{\figTick 0.4}}}
  \put(56.00,32.00){\makebox(0,0)[r]{{\figTick 0.5}}}

  \put(63.80,6.76){\makebox(0,0)[c]{{\figTick $10^2$}}}
  \put(76.80,6.76){\makebox(0,0)[c]{{\figTick $10^3$}}}
  \put(89.80,6.76){\makebox(0,0)[c]{{\figTick $10^4$}}}

  \put(58.70,9.90){\makebox(0,0)[l]{\textcolor{figTextGray}{\figNote small + low-contrast region}}}

  \put(87.55,30.98){\makebox(0,0)[l]{{\figLegend FN-dominated}}}
  \put(87.55,29.03){\makebox(0,0)[l]{{\figLegend FP-dominated}}}
  \put(87.55,27.08){\makebox(0,0)[l]{{\figLegend Boundary}}}
  \put(87.55,25.13){\makebox(0,0)[l]{{\figLegend Confusion}}}
\end{overpic}
\vspace{-1cm}
\caption{Error analysis of open-vocabulary mammographic lesion segmentation.  (a) Decomposition of pixel-level errors into false-negative, false-positive, boundary, and category-confusion errors.  (b) Failure distribution with respect to lesion size and local contrast.  Small and low-contrast lesions are the dominant source of remaining errors, and colors indicate the dominant error type for each lesion component.}
\label{fig:error_analysis}
\end{figure*}

\subsection{Qualitative Results}
\label{sec:qualitative_results}

\definecolor{figGT}{HTML}{F4B400}
\definecolor{figPred}{HTML}{20C6CF}
\definecolor{figTP}{HTML}{22C55E}
\definecolor{figFP}{HTML}{EF3340}
\definecolor{figFN}{HTML}{3367D6}
\definecolor{figUnc}{HTML}{7E57C2}
\definecolor{figSuccess}{HTML}{1DBF73}
\definecolor{figPartial}{HTML}{F39C12}
\definecolor{figFailure}{HTML}{EF3340}

\newcommand{\figRowTitle}{\fontsize{8}{9}\selectfont\bfseries}
\newcommand{\figDataset}{\fontsize{7.6}{8.4}\selectfont\bfseries}
\newcommand{\figLabel}{\fontsize{6.8}{7.5}\selectfont}
\newcommand{\figCaseText}{\fontsize{6.8}{7.5}\selectfont\bfseries}
\newcommand{\figBottomText}{\fontsize{6.8}{7.5}\selectfont}
\newcommand{\figLegendText}{\fontsize{6.2}{7}\selectfont}
\newcommand{\figcasebox}[2]{%
  \begingroup\setlength{\fboxsep}{0.8pt}%
  \colorbox{#1}{\textcolor{white}{\figCaseText #2}}%
  \endgroup}
\newcommand{\figlegenditem}[2]{%
  \textcolor{#1}{\rule{1.1ex}{1.1ex}}\hspace{0.35ex}#2}

\begin{figure*}[!t]
\centering
\begin{overpic}[width=\textwidth]{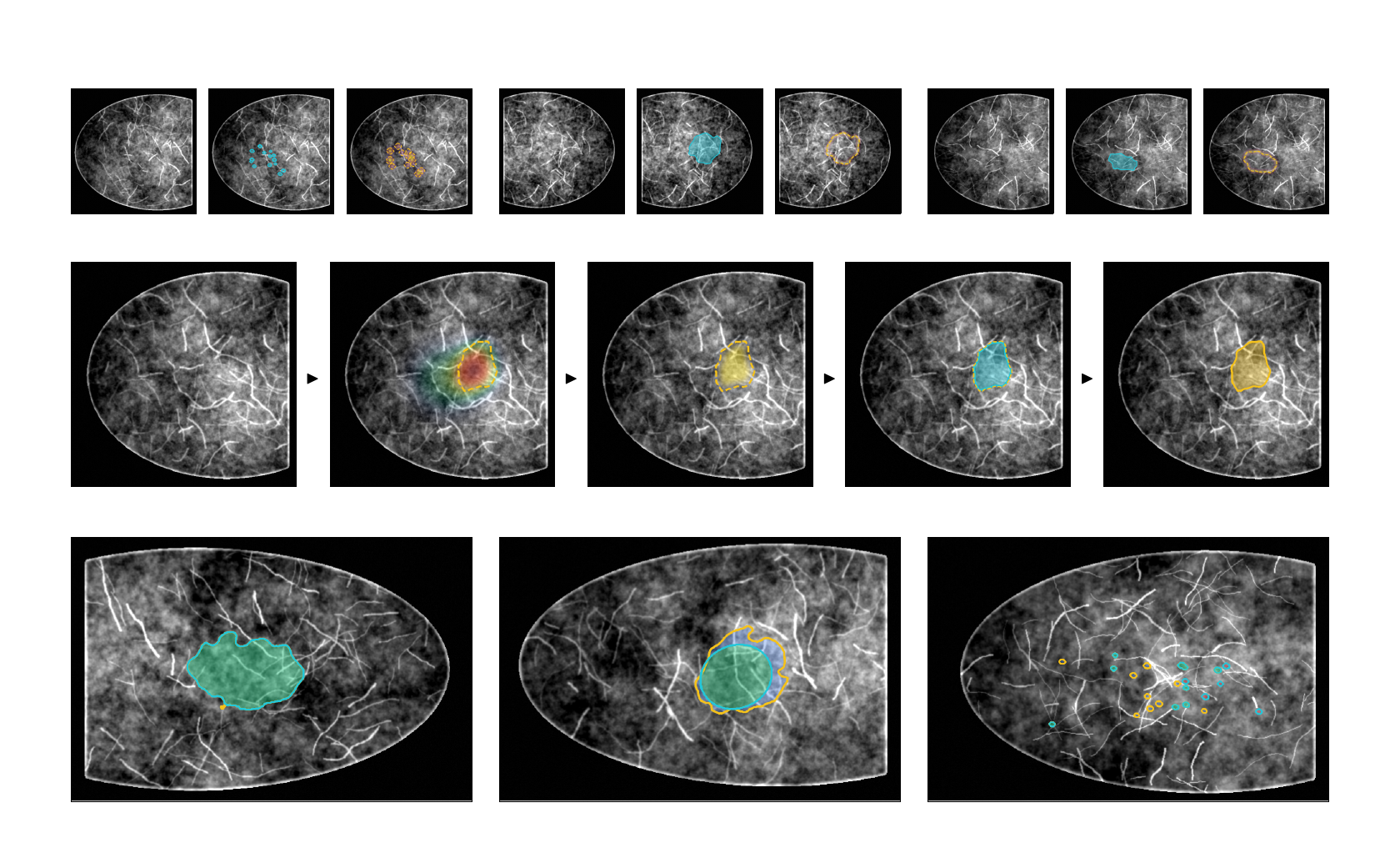}
  \put(2.56,57.70){\makebox(0,0)[l]{{\figRowTitle (a) Cross-dataset generalization}}}

  \put(21.55,56.13){\makebox(0,0)[c]{{\figDataset MIAS}}}
  \put(50.36,56.13){\makebox(0,0)[c]{{\figDataset INBreast}}}
  \put(78.99,56.13){\makebox(0,0)[c]{{\figDataset Mammo-Bench}}}

  \put(11.43,54.88){\makebox(0,0)[c]{{\figLabel Input}}}
  \put(20.83,54.88){\makebox(0,0)[c]{{\figLabel Prediction}}}
  \put(30.36,54.88){\makebox(0,0)[c]{{\figLabel Uncertainty}}}
  \put(40.48,54.88){\makebox(0,0)[c]{{\figLabel Input}}}
  \put(49.82,54.88){\makebox(0,0)[c]{{\figLabel Prediction}}}
  \put(59.35,54.88){\makebox(0,0)[c]{{\figLabel Uncertainty}}}
  \put(69.52,54.88){\makebox(0,0)[c]{{\figLabel Input}}}
  \put(78.87,54.88){\makebox(0,0)[c]{{\figLabel Prediction}}}
  \put(88.39,54.88){\makebox(0,0)[c]{{\figLabel Uncertainty}}}

  \put(2.56,43.75){\makebox(0,0)[l]{{\figRowTitle (b) From image-level attention to segmentation}}}

  \put(14.05,42.56){\makebox(0,0)[c]{{\figLabel Input}}}
  \put(32.20,42.56){\makebox(0,0)[c]{{\figLabel GAP CAM}}}
  \put(49.94,42.56){\makebox(0,0)[c]{{\figLabel LA-Map}}}
  \put(68.04,42.56){\makebox(0,0)[c]{{\figLabel Ours Mask}}}
  \put(86.07,42.56){\makebox(0,0)[c]{{\figLabel Ground Truth}}}

  \put(2.56,24.05){\makebox(0,0)[l]{{\figRowTitle (c) Reliability cases}}}

  \put(19.40,22.92){\makebox(0,0)[c]{\figcasebox{figSuccess}{Success}}}
  \put(49.94,22.92){\makebox(0,0)[c]{\figcasebox{figPartial}{Partial}}}
  \put(80.60,22.92){\makebox(0,0)[c]{\figcasebox{figFailure}{Failure}}}

  \put(19.40,2.56){\makebox(0,0)[c]{{\figBottomText compact lesion cluster}}}
  \put(49.88,2.56){\makebox(0,0)[c]{{\figBottomText fuzzy boundary}}}
  \put(80.83,2.56){\makebox(0,0)[c]{{\figBottomText extremely subtle finding}}}

  \put(50.00,1.13){\makebox(0,0)[c]{{\figLegendText
    \figlegenditem{figGT}{GT}\quad
    \figlegenditem{figPred}{Prediction}\quad
    \figlegenditem{figTP}{TP}\quad
    \figlegenditem{figFP}{FP}\quad
    \figlegenditem{figFN}{FN}\quad
    \figlegenditem{figUnc}{Uncertainty}}}}
\end{overpic}
\caption{Generalization and reliability visualization. (a) Cross-dataset qualitative results on MIAS, INBreast, and Mammo-Bench.  Each block shows the mammogram crop, prediction, and uncertainty map. (b) Visualization of the supervision transfer process from global image-level activation to lesion-wise attention and final segmentation.  Compared with the global average pooling CAM, the proposed LA-Map focuses more accurately on lesion-relevant regions. (c) Reliability analysis with representative successful, partial, and failure cases. Compact lesions are localized reliably, while fuzzy boundaries and extremely subtle calcifications remain difficult.}
\label{fig:generalization_reliability}
\end{figure*}

Fig.~\ref{fig:generalization_reliability} further analyzes cross-dataset generalization and prediction reliability. Panel~(a) shows qualitative results on MIAS, INBreast, and Mammo-Bench, demonstrating that the model maintains consistent lesion localization across different image resolutions, acquisition conditions, and annotation styles. Panel~(b) visualizes the transition from image-level attention to segmentation. Compared with the global average pooling CAM, the proposed LA-Map suppresses responses from normal tissue, concentrates on lesion-relevant pixels, and produces a cleaner final mask. Panel~(c) presents representative reliability cases at different difficulty levels. The method localizes compact lesion clusters reliably and can partially handle ambiguous boundaries, whereas fuzzy lesion margins and extremely subtle calcifications remain challenging.

\section{Conclusion}
\label{sec:conclusion}

In this paper, we study open-vocabulary semantic segmentation for mammography images by leveraging additional image-level classification datasets. We propose a unified framework that bridges the gap between image-level supervision and pixel-level segmentation. We construct a baseline that incorporates classification data to extend the visual vocabulary of segmentation models. Motivated by the observation that segmentation models can implicitly group unseen lesion pixels, we introduce a lesion-aware classifier to apply image-level supervision to semantically relevant regions. Extensive experiments demonstrate that the proposed framework consistently improves the baseline and achieves strong performance on both base and novel lesion categories. These results show that leveraging additional classification datasets is an effective strategy for improving mammography image segmentation.

\section*{Acknowledgment}

This work is supported by National Science Foundation of China (NSFC92470123, NSFC62276005) and the State Key Laboratory of General Artificial Intelligence.

\bibliographystyle{IEEEtran}
\bibliography{reference}

\end{document}